\documentclass[10pt,twocolumn,letterpaper]{article}

\usepackage[pagenumbers]{cvpr} 

\definecolor{cvprblue}{rgb}{0.21,0.49,0.74}
\usepackage[pagebackref,breaklinks,colorlinks,allcolors=cvprblue]{hyperref}

\newcommand{\modelname}{MS-Temba}

\newcommand{\block}{Temba}

\newcommand{\fuser}{Multi-scale Mamba Fuser}
\newcommand{\fuserabbr}{MS-Fuser}

\newcommand{\xmark}{\ding{55}}

\usepackage{amsmath}
\usepackage{amssymb}
\usepackage{booktabs}
\usepackage{tabularx}
\usepackage{colortbl}
\usepackage{pifont}
\usepackage{epigraph}
\usepackage{subcaption}
\usepackage{rotating}
\usepackage{multirow}
\usepackage{soul}
\usepackage{graphicx}
\usepackage{caption}
\usepackage{xcolor} 
\usepackage{tikz}

\usepackage{tcolorbox}
\usepackage[normalem]{ulem} 
\usepackage{calc} 

\definecolor{lightblue}{RGB}{173,216,230}
\definecolor{customred}{HTML}{C55A11} 

\definecolor{challenge1}{HTML}{C55A11} 
\definecolor{challenge21}{HTML}{0E8C0E} 
\definecolor{challenge22}{HTML}{DBBA58} 
\definecolor{challenge23}{HTML}{797979} 
\definecolor{challenge3}{HTML}{005493} 

\usepackage{stackengine}

\def\paperID{1461} 
\def\confName{CVPR}
\def\confYear{2026}

\title{
    \begin{minipage}{0.125\textwidth}
        \includegraphics[width=\textwidth]{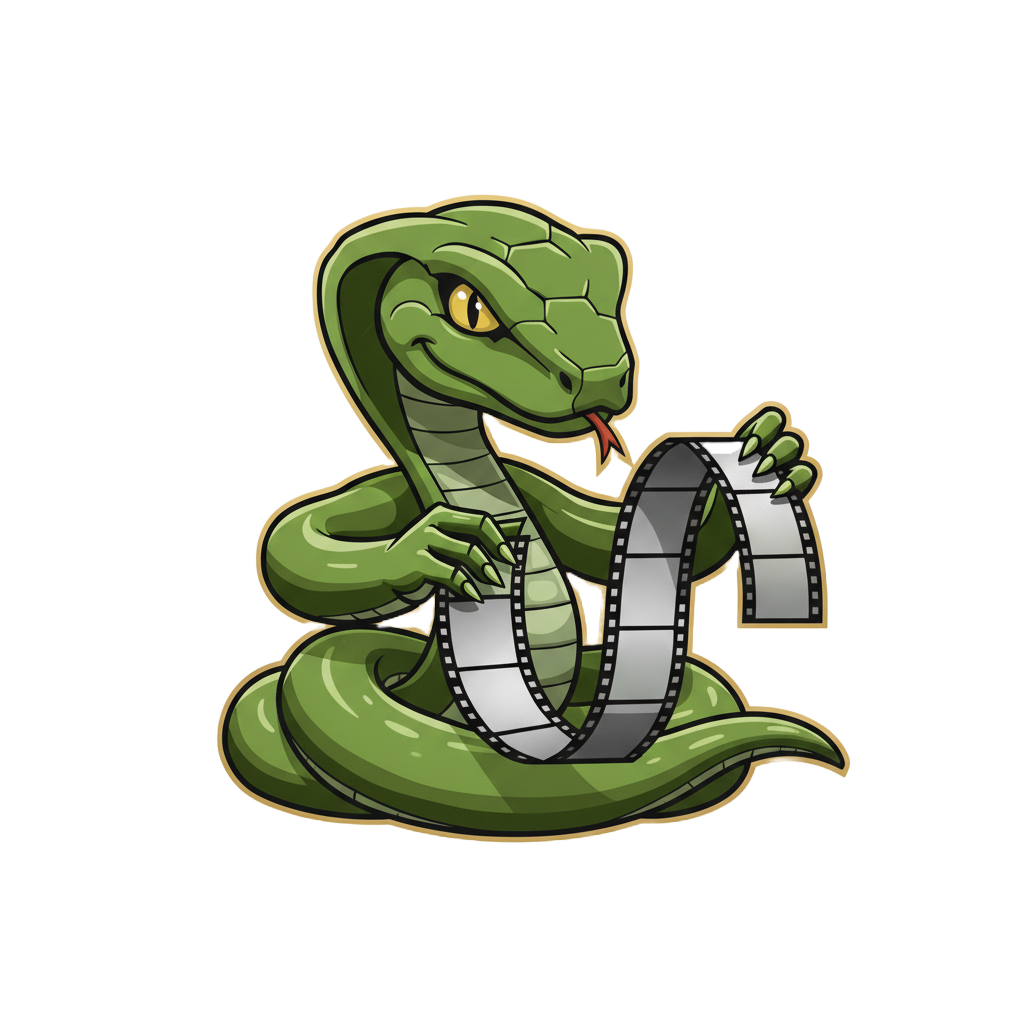}
    \end{minipage}
    \hspace{-0.16\textwidth}
    \begin{minipage}{0.85\textwidth}
        \centering
        \modelname: Multi-Scale Temporal Mamba for 
        \\ Understanding Long Untrimmed Videos
    \end{minipage}
}

\author{
\textbf{Arkaprava Sinha$^{1}$} \hspace*{0.5em}
\textbf{Monish Soundar Raj$^{2}$} \hspace*{0.5em} \\
\textbf{Pu Wang$^{1}$} \hspace*{0.5em}
\textbf{Ahmed Helmy$^{1}$} \hspace*{0.5em}
\textbf{Hieu Le$^{1}$} \hspace*{0.5em}
\textbf{Srijan Das$^{1}$} \hspace*{0.5em}
\vspace*{0.25em}
\\
$^{1}$ UNC Charlotte \hspace*{0.1em}
$^{2}$ UNC Chapel Hill \hspace*{0.1em}
\vspace*{0.25em}
\\
{\tt\normalsize \href{https://mstemba.github.io}{https://mstemba.github.io}}
}
\begin{document}

\twocolumn[{
\renewcommand\twocolumn[1][]{#1}%
\maketitle
    \vspace{-0.7cm}
    \centering
    \scalebox{0.98}{  
    \includegraphics[width=1\textwidth]{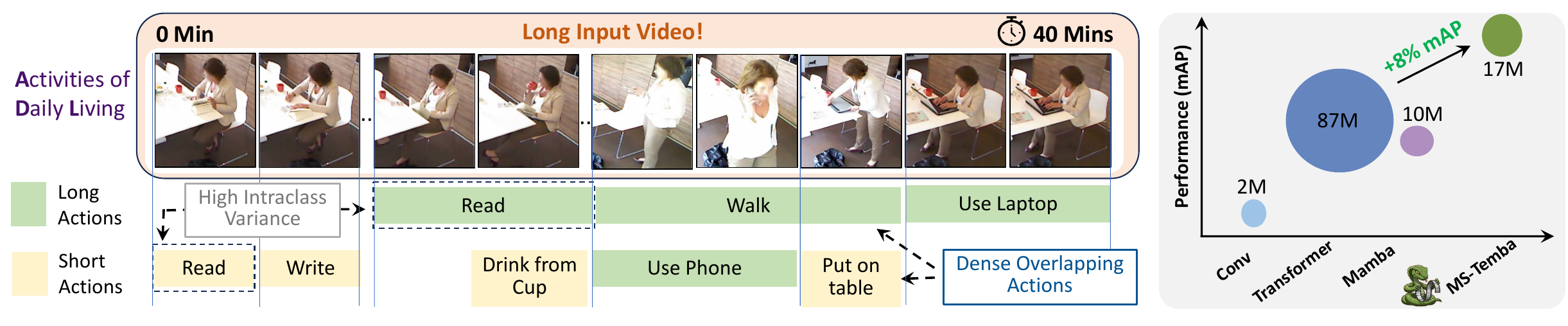}
    }
    \setcounter{figure}{0} 
    \captionsetup{hypcap=false, skip=4pt}
    \captionof{figure}{\textbf{Left:} Temporal Action Detection poses unique challenges, including the need for models capable of processing long video sequences, the presence of both short and long actions with high intra-class variance, and the complexity of densely overlapping actions. \textbf{Right:} Our proposed method, \textbf{\modelname}, achieves \textbf{state-of-the-art performance} while being \textbf{$\mathbf{5\times}$ more parameter-efficient} compared to transformer-based approaches. \vspace{0.5cm}}
    \label{fig:teaser}
}]

\begin{abstract}
Temporal Action Detection (TAD) in untrimmed videos poses significant challenges, particularly for Activities of Daily Living (ADL) requiring models to (1) process long-duration videos, (2) capture temporal variations in actions, and (3) simultaneously detect dense overlapping actions. Existing CNN and Transformer-based approaches, struggle to jointly capture fine-grained detail and long-range structure at scale.
State-space Model (SSM) based Mamba offers powerful long-range modeling, but na\"ive application to TAD collapses fine-grained temporal structure and fails to account for the challenges inherent to TAD.
To this end, we propose \textbf{Multi-Scale Temporal Mamba (\modelname)}, which extends Mamba to TAD with newly introduced \textit{dilated SSMs}. Each \block~block, comprising dilated SSMs coupled with our proposed additional losses, enables the learning of discriminative representations across temporal scales. A lightweight \textbf{\fuser} then unifies these multi-scale features via SSM-based aggregation, yielding precise action-boundary localization. 
With only 17M parameters, \modelname~achieves state-of-the-art performance on densely labeled ADL benchmarks TSU \& Charades, and further generalizes to long-form video summarization, setting new state-of-the-art results on TVSum \& SumMe. \vspace{-0.5cm}
\end{abstract}

\section{Introduction}
\label{sec:intro}

Understanding human actions in long, untrimmed videos is fundamentally harder than classifying short, curated clips~\cite{kinetics, feichtenhofer2017spatiotemporal, vificlip}. Real-world recordings, especially in Activities of Daily Living (ADL) settings, span hours, capturing people \textit{walking}, \textit{eating}, \textit{using phones}, often simultaneously and with no clear start or end boundaries. Within the same sequence, brief atomic events such as \textit{Drink from Cup} intertwine with prolonged activities like \textit{Use Laptop}, producing highly irregular, overlapping temporal patterns. 

Temporal Action Detection (TAD) aims to make sense of this continuous stream by recognizing actions and precisely localizing when they occur. This capability is crucial to ADL applications such as patient monitoring, assistive living, and smart homes~\cite{DML-smartactions, smarthome}, where systems must interpret complex, concurrent behaviors unfolding over long durations. 
As illustrated in \Cref{fig:teaser}, realistic ADL videos pose three key challenges: (\textbf{\textcolor{challenge1}{Challenge 1}}) capturing complex temporal dependencies in long, untrimmed videos ($\sim$ 40 min) (\textbf{\textcolor{challenge23}{Challenge 2}}) representing actions spanning vastly different temporal scales, ranging from short atomic events such as \textit{Drink from Cup} to long sustained activities like \textit{Use Laptop}, while handling intra-class temporal variations (\eg \textit{Read} occurring at different temporal scales), and (\textbf{\textcolor{challenge3}{Challenge 3}}) detecting densely overlapping actions, where multiple activities (\eg, \textit{Walk} and \textit{Use Phone}) occur simultaneously. Addressing these challenges requires models that can reason jointly over short and long range dependencies, disentangle concurrent actions and maintain precise temporal localization across complex ADL contexts. 

Existing TAD frameworks, typically built on temporal CNNs or Transformers, struggle to meet these demands. Temporal CNNs capture local temporal patterns more efficiently, but their limited receptive fields hinder modeling of long-range dependencies. Transformer-based architectures~\cite{MLAD, dai2022mstct} have achieved strong performance in TAD by leveraging global self-attention, yet their quadratic complexity makes them less scalable to long video sequences. Recently, Mamba architectures based on ``\textit{state-space models}" (SSMs)~\cite{mamba, s4} have emerged as a promising alternative, offering linear-time information propagation across temporal sequences and enabling efficient long-range reasoning in video understanding \cite{vim, videomamba, chen2024video}. However, existing SSMs~\cite{videomamba, manta, videomambapro} typically operate at a single temporal scale, often blending short-term motion with surrounding context and degrading fine temporal granularity. This hinders their adaptation to TAD, particularly in densely overlapping ADL scenarios.

\begin{figure}
    \centering
    \scalebox{0.257}{
    \includegraphics{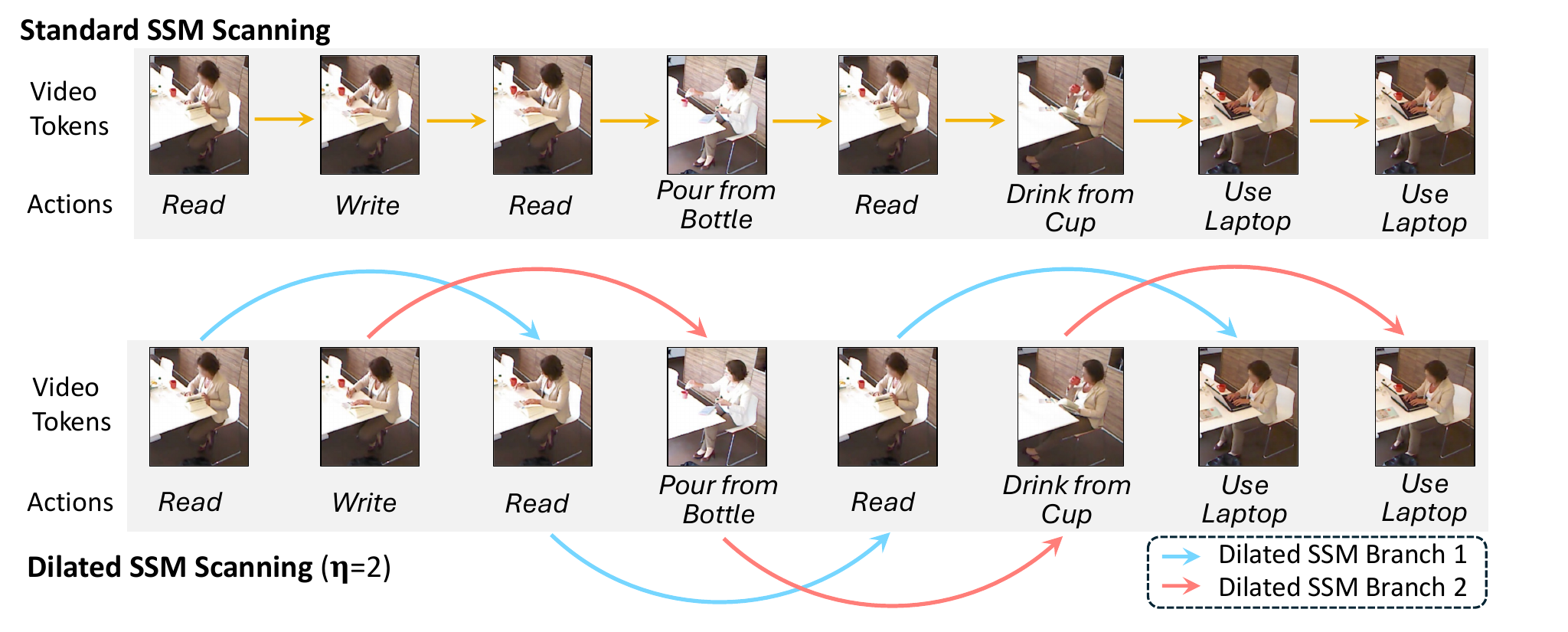}}
    \caption{Comparison between standard and dilated SSM scanning}
    \label{fig:scanning}
    \vspace{-6mm}
\end{figure}

To this end, we propose \textbf{Multi-Scale Temporal Mamba (\modelname)}, which extends the state-space formulation into a novel \textit{multi-scale}, \textit{dilated} architecture designed to capture both short- and long-term temporal dynamics within a unified framework. 
The core of \modelname~lies in its \textbf{\block~Blocks}, introducing \textit{dilated SSMs}. The key intuition is straightforward: rather than processing the video sequence with a single, fixed receptive field, we employ multiple dilated SSM branches, each operating at a distinct temporal stride. This design complements the standard SSM scanning process, which traverses all tokens sequentially (\Cref{fig:scanning}). Short-dilation branches focus on transient, fine-grained actions, whereas long-dilation branches capture extended temporal dependencies. 
To ensure coherence across these asynchronous branches, we introduce a projection-based alignment that harmonizes their action predictions and mitigates scale drift. In addition, we employ scale-aware auxiliary supervision that encourages blocks to learn scale-specific, temporally discriminative representations.

Furthermore, we design a lightweight \textbf{\fuser~(\fuserabbr)} that integrates the complementary representations learned by \block~blocks through SSM-based aggregation, producing temporally coherent and boundary-accurate predictions. This state-space formulation preserves the efficiency and modeling strength of Mamba while restoring the temporal sensitivity necessary for dense, overlapping ADL scenarios. 
Together, \block~and \fuserabbr~enable \modelname~to jointly model fine-grained short-term dynamics and extended long-range structures, offering a comprehensive solution for robust temporal action detection in complex video sequences. To assess the generalizability of \modelname~beyond TAD, we evaluate it on video summarization, the task of condensing long videos into shorter, semantically meaningful sequences that preserve the overall narrative. \modelname~exhibits superior performance over competitive baselines, indicating its capacity to generalize to long-form video understanding tasks.

To summarize the main contributions:
\begin{itemize}
    \item We propose \textbf{Multi-Scale Temporal Mamba} (\modelname), the first Mamba-based architecture tailored to densely labeled TAD in untrimmed videos.
    \item We introduce \textbf{Dilated SSMs} for multi-scale temporal representation learning and design a lightweight \textbf{\fuser} (\fuserabbr) module that aggregates these complementary scales via SSM-based fusion.
    \item \modelname~achieves state-of-the-art performance on densely labeled TAD benchmarks, TSU and Charades, with only 17M parameters, and further demonstrates strong generalization attaining state-of-the-art results on video summarization benchmarks (TVSum \& SumMe).
\end{itemize}

\section{Related Work}
\label{sec:relatedw}

\textbf{Temporal Action Detection.} Understanding actions in long temporal sequences has become a key task in computer vision~\cite{gleason2019proposal, lin2017single, gtad, Damen2018EPICKITCHENS, dai2019tan, zhao2019hacs}. Early action detection approaches focused on videos with sparse action labels~\cite{THUMOS14, zhao2019hacs, caba2015activitynet}, primarily using proposal-based methods~\cite{RC3d, SSN} inspired by object detection to generate action proposals. However, temporal action detection with densely labeled action distribution offers greater challenges due to concurrent actions, where proposal-based methods struggle with the combinatorial explosion in proposal generation. End-to-end methods~\cite{pointtad, freetad} jointly train the visual backbone and perform temporal modeling while optimizing the backbone for action detection. However, these methods are computationally prohibitive and forces the video to be broken into short windows, losing global temporal context.

This led to the adoption of temporal convolution-based models~\cite{superevent, mstcn, bsn, bmn} for processing long videos. Convolutional methods like PDAN~\cite{PDAN} and TGM~\cite{TGM1} use specialized kernels designed to model long-term dependencies and capture composite actions. However, the shared kernels in temporal convolutions are limited to neighboring frames, and even with dilations, struggle to capture global relationships between frames. This limitation is significant for densely labeled Temporal Action Detection, where long-term actions are common. In response, transformer-based architectures~\cite{attention, dosovitskiy2020vit, deit, liu2021swin, mvit1}, which have proven effective in computer vision, have been adapted for action detection~\cite{MLAD, dai2022mstct, asl, pointtad, haan, ryoo2023token}. These architectures can model both short- and long-term dependencies in videos. For example, MLAD~\cite{MLAD} employs a Transformer encoder to capture cross-class features and temporal relationships, while MS-TCT~\cite{dai2022mstct} uses a temporal hierarchy to model local relationships through temporal convolutions and global interactions through attention. These transformer based approaches enhances multi-scale representation learning, but the effectiveness comes at a high computational cost. The linear complexity of state space models~\cite{mamba} makes it an ideal test bed for processing long untrimmed videos. 

\noindent
\textbf{State Space Models (SSMs).} Recently, Structured State Space Model (S4)~\cite{s4} has gained popularity in effectively processing long sequence lengths in linear complexity. Mamba~\cite{mamba} addresses some challenges of S4 (selective copying and induction heads) and introduces data-dependent SSM layer and a hardware aware efficient algorithm. 
Consequently, Mamba has been adapted for various computer vision downstream tasks using static images such as image classification, object detection, semantic segmentation, image restoration in~\cite{vim, guo2024mambair, he2025pan, xing2024segmamba, liu2024swin, archit2024vim, vmamba}.
Naturally, these state space models have been adapted for videos in a wide range of applications such as action classification~\cite{videomamba, videomambapro, zubic2024state}, video object segmentation~\cite{ yang2024vivim}, video retrieval~\cite{tang2024muse}, action anticipation~\cite{manta}, etc.
However, these methods are trained on videos which span only few seconds and have a global class per video. It is non-trivial to adapt these methods for action detection since action detectors requires two stage training, first feature extraction from untrimmed videos using a visual backbone and then temporal learning of these features for dense action-label prediction. 
For long-range temporal modeling, TranS4mer~\cite{TranS4mer} has shown effectiveness in video classification and movie scene detection tasks by identifying scene changes within videos. MOGO~\cite{mogo} performs video action detection by spatially localizing and classifying atomic actions in videos.  Closest to our approach, Video Mamba Suite~\cite{chen2024video} performs action localization on datasets like HACS~\cite{zhao2019hacs} and GTEA~\cite{gtea}. These Mamba-based methods rely on frame sampling and compressing all information within their architectures limiting them to process videos around 3 minutes.
However, in this work, we target untrimmed videos exceeding 40 minutes, featuring dense labels and co-occurring actions.
State space models have not yet been applied to action detection in such untrimmed videos. To the best of our knowledge, \modelname~is the first to leverage Mamba for the task of temporal action detection in densely labeled untrimmed videos. 
\vspace{-2.5mm}
\section{Preliminaries}
\label{sec:prelims}
State space models map continuous input sequences $x(t) \in \mathbb{R}$ to continuous outputs $y(t) \in \mathbb{R}$ using a hidden state $h(t) \in \mathbb{R}^n$. For the above mapping and updating of the hidden state, SSMs use a transition state matrix $\mathbf{A} \in \mathbb{R}^{n \times n}$ projection matrices $\mathbf{B}\in \mathbb{R}^{n \times 1}$ and $\mathbf{C}\in \mathbb{R}^{1 \times n}$. The update equations are given by 
\begin{equation}
    \begin{aligned}
        h'(t) &= \mathbf{A}h(t) + \mathbf{B}x(t) \\
        y(t) &= \mathbf{C}h(t)
    \end{aligned}
\end{equation}
Previous works ~\cite{s4, mamba, vim}, use discrete versions of the above system using a discretization parameter $\mathbf{\Delta}$ known as the step size. Following is an example of a discretization transformation using $\mathbf{\Delta}$, known as zero-order hold (ZOH): 
\begin{equation}
  \begin{aligned}
    \overline{\mathbf{A}} &= \exp(\mathbf{\Delta}\mathbf{A}) \\
    \overline{\mathbf{B}} &= (\mathbf{\Delta}\mathbf{A})^{-1}  (\exp(\mathbf{\Delta}\mathbf{A}) - \mathbf{I})\mathbf{\Delta}\mathbf{B}
\end{aligned}  
\end{equation}

This discretization of $\mathbf{A}$, $\mathbf{B}$ to $\overline{\mathbf{A}}$ and $\overline{\mathbf{B}}$ leads to the following updated state-space equations:
\begin{equation}
   \begin{aligned}
    h_t &= \overline{\mathbf{A}}h_{t-1} + \overline{\mathbf{B}}x_t \\
    y_t &= \mathbf{C}h_t
\end{aligned} 
\end{equation}

Standard SSMs are typically designed to process 1D inputs in a linear sequence. However, \cite{vim, videomamba} introduce bidirectional scanning of tokens, which are more effective for vision tasks. In this approach, the SSM operates on the input sequence $v$ and it's reversed counterpart $v'$, fusing the projected outputs for robust bidirectional sequence modeling. 
This enhancement enables more effective representation learning, especially in domains requiring complex temporal dependencies. In this work, we extend the Mamba based architectures for understanding long untrimmed videos.
\begin{figure}
    \centering
    \scalebox{0.41}{
    \includegraphics{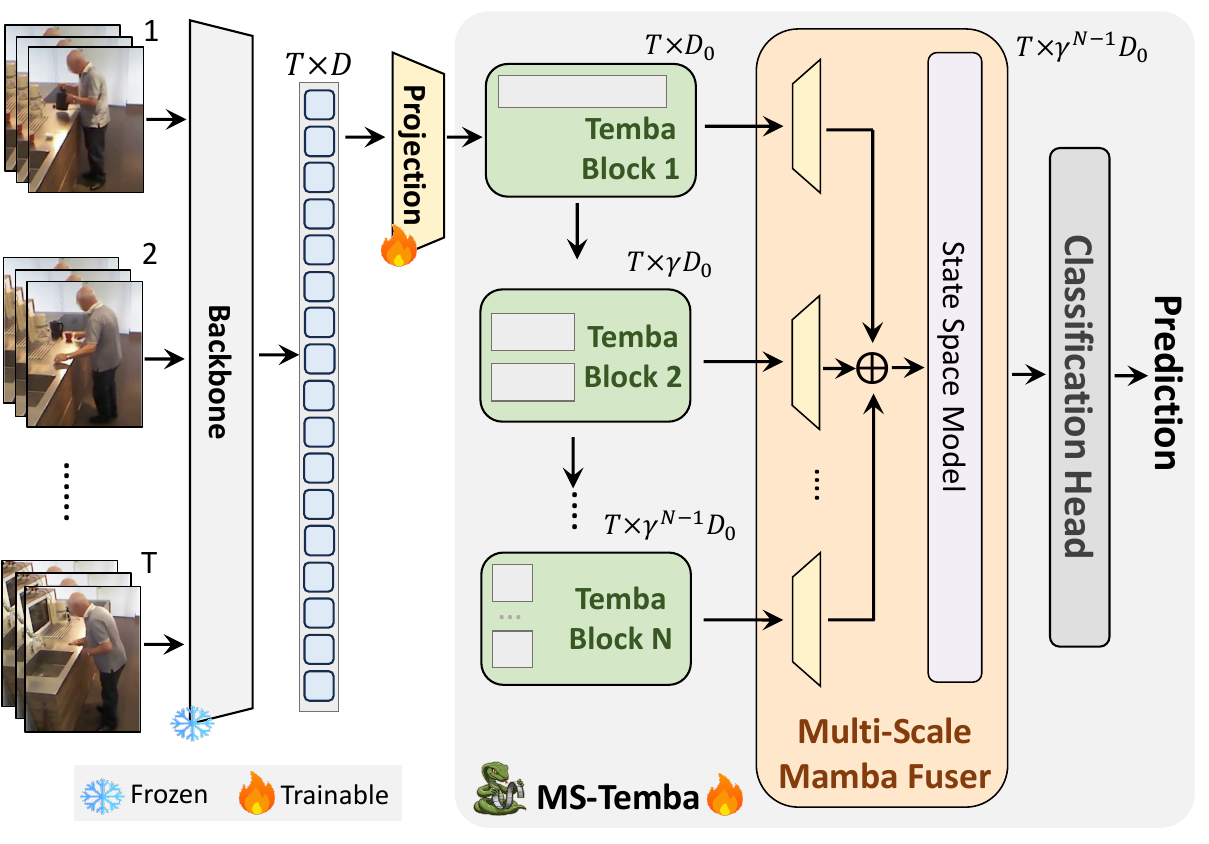}}
    \caption{Overall Architecture of \textbf{Multiscale Temporal Mamba (\modelname)} for action detection. \modelname~is composed of a frozen pretrained Visual Backbone, Temporal Mamba (\block) Blocks for learning representations at multiple temporal scale spaces through dilated SSMs. The \fuser~employs an SSM for effectively fusing the multi-scale features which is projected by the Classification Head for dense action detection.}
    \label{fig:overall}
    \vspace{-6mm}
\end{figure}
\vspace{-2.5mm}
\section{Proposed \modelname}
\label{sec:method}
In Temporal Action Detection (TAD), we aim to classify multiple activities at each time-step of an untrimmed video. Given a video of $T$ temporal segments, we associate an action label $\mathrm{Y} = \{y_t\}$ for each timestep $t$. 

 We introduce \textbf{Temporal Mamba (\modelname)}, a TAD model capable of detecting long as well as short actions efficiently in long untrimmed videos. The overall architecture of \modelname~is illustrated in \Cref{fig:overall}. Our proposed \modelname~consists of: i) a \textbf{Visual Backbone} to extract discriminative features from untrimmed videos, ii) \textbf{Temba} blocks which capture temporal information at different scales, iii) \textbf{\fuser} to combine the learned temporal representations across different temporal scales, and iv) a \textbf{Classification Head} to predict multiple actions within each temporal segment of the input video. 
In the following sections, we detail each of these components.

\subsection{Visual Backbone}
The input to \modelname~is an untrimmed video potentially spanning long durations (several minutes). Training an end-to-end model across both spatial and temporal dimensions poses significant computational demands. Consequently, in line with prior work~\cite{dai2022mstct, PDAN, mstcn, zhangtqn}, we first segment the untrimmed video into $T$ contiguous, non-overlapping segments, each consisting of 8 or 16 frames. Each segment, known as a visual token, is encoded via a visual backbone, capturing spatial features through either a 3D ConvNet (e.g., I3D~\cite{i3d}) pre-trained on the target dataset, or a larger foundational model such as CLIP~\cite{CLIP}. Unlike action classification tasks, sparse frame sampling is unsuitable here, as it risks omitting short-duration actions crucial for detection. The segment-wise representations are then stacked to form an input feature vector $v \in \mathbb{R}^{B \times T \times D}$, where $B$ is the batch of videos, $D$ is the feature dimension and $T$ is the total temporal sequence length. Finally, $v$ is projected to $z_0 \in \mathbb{R}^{B \times T \times D_0}$ to obtain a latent representation of the input, which serves as input tokens to our proposed Temba Blocks.

\begin{figure}
    \centering
    \scalebox{0.35}{
    \includegraphics{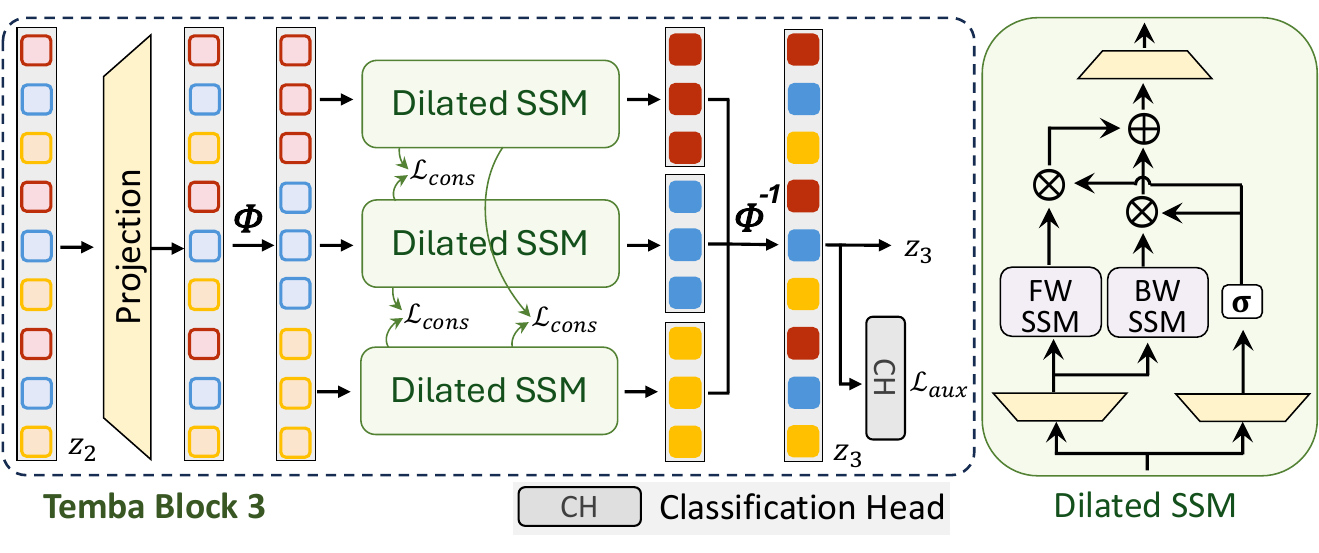}}
    \caption{\block~Block 3 with $\eta=3$. Tokens are linearly projected, grouped with stride $\eta$ using $\Phi$, processed by SSMs, aligned with $\mathcal{L}_{\text{cons}}$, and reassembled by $\Phi^{-1}$. A classification head then produces the auxiliary loss $\mathcal{L}_{\text{aux}}$.}
    \label{fig:temba}
    \vspace{-6mm}
\end{figure}

\subsection{Temporal Mamba (Temba) Blocks}
TAD requires models that simultaneously capture fine-grained dynamics and long-range temporal dependencies. We introduce \block~blocks, the core computational unit of \modelname, which is a novel formulation of \textit{dilated State Space Models} for multi-scale temporal representation learning. Each \block~block consists of two key components: (i) a set of \textit{dilated SSM branches}, (ii) \textit{scale-aware auxiliary supervision}.

Each Temba block begins with a linear projection that increases the representational capacity, enabling deeper temporal abstractions as the hierarchy progresses. The feature dimensionality increases by a factor of $\gamma$ relative to the preceding block. Given the weight matrix $\mathbf{W}_k$ and bias term $\boldsymbol{\beta}_k$ for the $k$-th block, the transformation is defined as:
\begin{equation}
    x_k = \mathbf{W}_k z_{k-1} + \boldsymbol{\beta}_k, \quad x_k \in \mathbb{R}^{B \times T \times \gamma D_{k-1}}, \quad k \geq 1.
\end{equation}
This progressive expansion enables each Temba block to model increasingly complex temporal dependencies while maintaining computational efficiency. In \Cref{fig:temba}, we illustrate the third \block~block with $\eta=3$.

\noindent\textbf{i) \textit{Dilated SSM}.} Each \block~consists of \textit{dilated SSM}, which allows \modelname~to capture dependencies at varying temporal strides. 
Given an input feature sequence $x_k \in \mathbb{R}^{B \times T \times D_k}$, we define a non-parametric, invertible mapping
$\Phi_{\eta}: \mathbb{R}^{B \times T \times D_k} \rightarrow \mathbb{R}^{\eta B \times \left\lceil T / \eta \right\rceil \times D_k}$,
which partitions each temporal sequence into \(\eta\) disjoint subsequences sampled with stride \(\eta\). The dilation rate \(\eta\) controls the temporal resolution at which each subset of tokens is processed. Concretely, letting
\begin{align}
\big(X^{(0)}, X^{(1)}, \dots, X^{(\eta-1)}\big) = \Phi_{\eta}(x),
\end{align}
each subset correspond to a dilated slice of the temporal axis:
\begin{equation} \begin{aligned} 
X^{(i)} &= \{\, x_t \mid t = i + j\eta,\; 0 \le j < \lceil T / \eta \rceil \,\},\\ &\quad i = 1, 2, \dots, \eta. \end{aligned} \end{equation}
In this formulation, $X^{(1)}$ contains tokens sampled at positions $1, 1+\eta, 1+2\eta, \dots$, while $X^{(2)}$ starts from the second token and follows the same stride. As a result, $\Phi_{\eta}$ divides the sequence into $\eta$ temporally distinct streams, each representing a different dilated phase of the input sequence.

Each subsequence is then processed by an individual dilated SSM branch parameterized by $(\mathbf{A}^{(i)}, \mathbf{B}^{(i)}, \mathbf{C}^{(i)})$. The output of the $i$-th SSM is expressed as
\begin{align}
    y^{(i)}_{t} &= \mathbf{C}^{(i)} h^{(i)}_{t}, \quad \mathrm{where} \\
        h^{(i)}_{t} &= \overline{\mathbf{A}^{(i)}} h^{(i)}_{t-1} + \overline{\mathbf{B}^{(i)}} X^{(i)}_{t}
\end{align}
Distinct parameterizations across the $\eta$ SSM branches allow each to learn temporal relationships at a unique dilation, resulting in diverse and complementary receptive fields. 

However, to ensure coherence across neighboring temporal contexts, we introduce a \textbf{projection alignment} mechanism that enforces consistency among the output projections of dilated SSM branches within each block. Intuitively, if an action is activated in the state representation of one SSM at time $t_s$, adjacent temporal segments $t_{s-1}$ and $t_{s+1}$ should exhibit similar activation patterns in adjacent dilated SSM branches in each block. 
To achieve this inductive bias, we define a \textbf{pairwise consistency loss} that aligns the projection matrices across the branches. 
Let $\mathbf{C}_i$ and $\mathbf{C}_j$ denote the output projection matrices of the $i$-th and $j$-th dilated SSMs in a block. The pairwise consistency loss for a pair \((i,j)\) is
\begin{equation}
    \mathcal{L}_{\text{cons}} = 1 - \text{sim}(\widehat{\mathbf{C}}_i, \widehat{\mathbf{C}}_j), \quad i \neq j,
\end{equation}
where $\widehat{\mathbf{C}}_i$, $\widehat{\mathbf{C}}_j$ are their fattened, \(\ell_2\)-normalized forms, and $\text{sim}(\cdot,\cdot)$ denotes cosine similarity. This pairwise consistency loss encourages the learned state 
projections to remain semantically aligned across SSMs within the same dilated block, while maintaining diversity across temporal receptive fields.

After the dilated SSM branches process their respective subsequences, the outputs are reassembled to restore the original temporal ordering via the inverse mapping:
\begin{equation}
    \Phi^{-1}_{\eta}: \mathbb{R}^{\eta B \times  \lceil T / \eta \rceil \times D_k} \rightarrow \mathbb{R}^{B \times T \times D_k}.
\end{equation}
The final sequence is then given by
\begin{align}
    z_k &= \Phi^{-1}_{\eta}(y^{(1)}, y^{(2)}, \dots, y^{(\eta)}) 
\end{align}
where $\Phi^{-1}_{\eta}(\Phi_{\eta}(x)) = x$ ensures that the rearrangement is bijective and preserves temporal fidelity. The reconstructed representation $z_k$ is then passed to the subsequent block, enabling hierarchical refinement over progressively increasing dilation rates.

\noindent \textbf{ii) \textit{Scale-Aware Auxiliary Supervision}.}
To further enrich the temporal abstraction, we stack multiple \textbf{Temba blocks} with progressively increasing dilation rates, where the $k$-th block uses $\eta = k$. Each block refines the representation obtained from the previous one:
\begin{equation}
    z_{k} = \mathrm{Temba}_{\eta=k}(z_{k-1}); \quad k \geq 1
\end{equation}
enabling hierarchical modeling across temporal scales. 
Blocks with smaller dilation factors capture local and short-term variations, while those with larger dilation factors integrate long-range dependencies. Through this hierarchical stacking, the model gradually builds a comprehensive temporal understanding that spans multiple scales, enabling accurate and efficient reasoning over long untrimmed videos.
To enhance discriminability and encourage scale-specific specialization, we employ a \textbf{scale-aware auxiliary supervision} mechanism. 
Each block is equipped with a lightweight classification head that produces block-level predictions $\mathrm{\hat{Y}}^k$ from its output representation. The auxiliary loss for each block is defined as a standard Binary Cross Entropy ($\mathrm{BCE}$) loss between its block-level predictions $\mathrm{\hat{Y}}^k$ and the ground-truth labels $\mathrm{Y}$:
\begin{equation}
    \mathcal{L}_{\text{aux}} = \mathrm{BCE}(\hat{\mathrm{Y}}^k, \mathrm{Y})
\end{equation}
ensuring that each block receives direct supervisory feedback. This auxiliary supervision constrains the temporal semantics of intermediate representations, maintaining each block’s inductive bias toward its respective receptive field and dilation scale. By promoting such scale-aware discriminability, \modelname~learns a hierarchy of temporally specialized features, thereby motivating the need for a dedicated \textit{fusion module} to effectively integrate these multi-scale representations into a unified temporal encoding. \vspace{-1mm}

\subsection{\fuser}

The \textbf{\fuser~(\fuserabbr)} aggregates the temporal representations at multiple temporal scales learned by the stacked Temba blocks to form a unified, scale-aware temporal embedding. Let $\{z_1, z_2, \dots, z_K\}$ denote the outputs of the $K$ Temba blocks, where each $z_k \in \mathbb{R}^{B \times T \times D_k}$ corresponds to the representation at dilation rate $\eta = k$. Since the feature dimensions $D_k$ may vary, each $z_k$ is first projected to a common embedding dimension $E$ through a linear transformation with weight matrix $\mathbf{W}^f_k$ and bias $\boldsymbol{\beta}^f_k$ for block $k$. The projected features are then fused to obtain a unified representation $z^f$:
\begin{align}
    \tilde{z}_k &= \mathbf{W}^f_k z_k + \boldsymbol{\beta}^f_k, \quad \tilde{z}_k \in \mathbb{R}^{B \times T \times E}. \\
    z^f &= \sum_{k=1}^{K} \tilde{z}_k
\end{align}
To capture cross-scale temporal dependencies, an additional SSM refines the fused representation through hidden state transitions:
\begin{align}
    h^{f}_t &= \overline{\mathbf{A}^{f}} h^{f}_{t-1} + \overline{\mathbf{B}^{f}} z^{f}_t, \\
    y^{f}_t &= \mathbf{C}^{f} h^{f}_t.
\end{align}
The resulting sequence $\{y^{f}_t\}$ encodes both fine-grained and long-range temporal relationships, allowing \fuserabbr~to effectively integrate complementary information from different dilation rates. This design ensures that earlier Temba blocks contribute detailed short-term cues, while later blocks provide coarse, long-horizon context, yielding a compact yet expressive multi-scale temporal representation well-suited for action detection in long untrimmed videos. Finally, the fused output ${y^{f}_t}$ is passed through a classification head to obtain multiclass predictions $\hat{\mathrm{Y}} = \{\hat{y}_t\}$ for each temporal segment for temporal action localization.\vspace{-1mm}

\subsection{Training \modelname}

For multi-label action detection, following prior works~\cite{dai2022mstct,MLAD}, we optimize the network using a BCE loss:
\begin{equation}
    \mathcal{L}_{\text{BCE}} = \mathrm{BCE}(\hat{\mathrm{Y}}, \mathrm{Y})
\end{equation}
where $\mathrm{Y}$ denotes the ground-truth label, and $\hat{\mathrm{Y}}$ is the model prediction.
The overall training objective combines the primary BCE loss, the pairwise consistency loss and the averaged auxiliary losses:
\begin{equation}
    \mathcal{L} = \mathcal{L}_{\text{BCE}} + \alpha\mathcal{L}_{\text{cons}} + \frac{\beta}{K}\sum_{k=1}^{K}\mathcal{L}_{\text{aux}},
\end{equation}
where $\alpha$ and $\beta$ controls the strength of the consistency and auxiliary losses respectively.  This composite objective enforces global discriminability, local temporal coherence, and scale-aware specialization, enabling robust temporal action detection in long untrimmed videos.
\begin{table}[h]
\centering
\caption{\textbf{Comparison with the State-of-the-art}. \textbf{Bold} denotes best scores, \underline{underline} denotes the second best.}
\setlength{\tabcolsep}{3pt} 
\scalebox{0.9}{ 
\begin{tabular}{l@{\hspace{3pt}}|c@{\hspace{3pt}}c@{\hspace{3pt}}|c@{\hspace{3pt}}c@{\hspace{3pt}}c}
\toprule
\textbf{Method} & \textbf{Visual} & \textbf{\#Param}  & \textbf{TSU}  & \textbf{Charades} \\
& \textbf{Backbone} & \textbf{(M)} & \textbf{(mAP)} & \textbf{(mAP)} \\
\midrule
Super-event~\cite{superevent} & I3D & 26  & 17.2 & 18.6  \\
TGM~\cite{TGM1} & I3D & 2  & 26.7  & 20.6  \\
PDAN~\cite{PDAN} & I3D & 6  & 32.7  & 23.7  \\
Coarse-Fine~\cite{kahatapitiya2021coarse} & X3D & 8  & - & 25.1 \\
MLAD~\cite{MLAD} & I3D & 21  & -  & 18.4  \\
PointTAD~\cite{pointtad} & I3D & -  & - & 22.1  \\
HAAN~\cite{haan} & I3D & - & -  & 25.2  \\
CTRN~\cite{dai2021ctrn} & I3D & 11  & 33.5 & \underline{25.3} \\
MS-TCT~\cite{dai2022mstct} & I3D & 87 & 33.7 & \textbf{25.4}  \\
DualDETR~\cite{dual_detr} & I3D & 21  & \underline{34.8} & 23.2 \\
\rowcolor{lightblue} \textbf{\modelname~(ours)} & I3D & 17 & \textbf{36.1} & \textbf{25.4} \\
\arrayrulecolor{black}\midrule
TTM~\cite{ryoo2023token} & ViViT & 89  & -  & 26.3  \\
AAN~\cite{dai2023aan} & CLIP & -   & 41.3  & 32.0  \\
MS-TCT~\cite{dai2022mstct} & CLIP & 87  & 40.6 & 31.9  \\
\rowcolor{lightblue} \textbf{\modelname~(ours)} & CLIP & 17  & \textbf{44.0} & \textbf{33.6} \\
\bottomrule
\end{tabular}}
\label{tab:sota}
\end{table}
\begin{table*}[h!]
\centering
\caption{\textbf{Performance on Action Conditioned Metrics on TSU.} $P_{AC}$ - Action Conditioned Precision; $R_{AC}$ - Action Conditioned Recall; $F1_{AC}$ - Action Conditioned F1-score; $mAP_{AC}$ - Action Conditioned Mean Average Precision. $\tau$ indicates the temporal window size.}
\scalebox{0.7}{
\begin{tabular}{l|cccc|cccc|cccc}
\toprule
\multirow{2}{*}{\textbf{Method}} & \multicolumn{4}{c|}{\boldmath$\tau = 0$} & \multicolumn{4}{c|}{\boldmath$\tau = 20$} & \multicolumn{4}{c}{\boldmath$\tau = 40$} \\
 & $\mathbf{P}_{AC}$ & $\mathbf{R}_{AC}$ & $\mathbf{F1}_{AC}$ & $\mathbf{mAP}_{AC}$  & $\mathbf{P}_{AC}$ & $\mathbf{R}_{AC}$ & $\mathbf{F1}_{AC}$ & $\mathbf{mAP}_{AC}$  & $\mathbf{P}_{AC}$ & $\mathbf{R}_{AC}$ & $\mathbf{F1}_{AC}$ & $\mathbf{mAP}_{AC}$ \\
\midrule
I3D~\cite{i3d} & 12.84 & 3.73 & 5.79 & 13.94  & 24.10 & 6.14  & 9.79 & 21.39 & 25.05 & 6.12 & 9.84 & 20.41 \\
I3D + MS-TCT~\cite{dai2022mstct} & 23.26 & 18.15 & 20.40 & 25.74  & 34.19 & 22.67  & 27.26 & 37.12 & 35.49 & 23.86 & 28.53 & 36.35\\
\rowcolor{lightblue}
I3D + \modelname~(ours)  & \textbf{25.68} & \textbf{19.90} & \textbf{22.43} & \textbf{27.03} & \textbf{38.13} & \textbf{24.59} & \textbf{29.90} & \textbf{38.86} & \textbf{38.96} & \textbf{24.84} & \textbf{30.34} & \textbf{38.05} \\
\midrule
CLIP + MS-TCT~\cite{dai2022mstct} & 26.49 & 21.34 & 23.64 & 31.14 & 40.78 & 30.93 & 35.18 & 43.44 & 42.66 & 32.40 & 36.80 & 43.36 \\
\rowcolor{lightblue}
CLIP + \modelname~(ours)  & \textbf{28.61} & \textbf{23.66} & \textbf{25.90} & \textbf{31.50} & \textbf{46.31} & \textbf{32.89} & \textbf{38.47} & \textbf{45.95} & \textbf{48.26} & \textbf{34.28} & \textbf{40.09} & \textbf{45.82} \\ 
\bottomrule
\end{tabular}}
\vspace{-3mm}
\label{tab:acmetrics}
\end{table*}
\begin{table*}[t]
\centering
\begin{minipage}[t]{0.28\textwidth}
  \vspace{-3pt}
  \vspace{\abovecaptionskip}%
  \centering
 \includegraphics[width=0.9\linewidth, height=0.12\textheight]{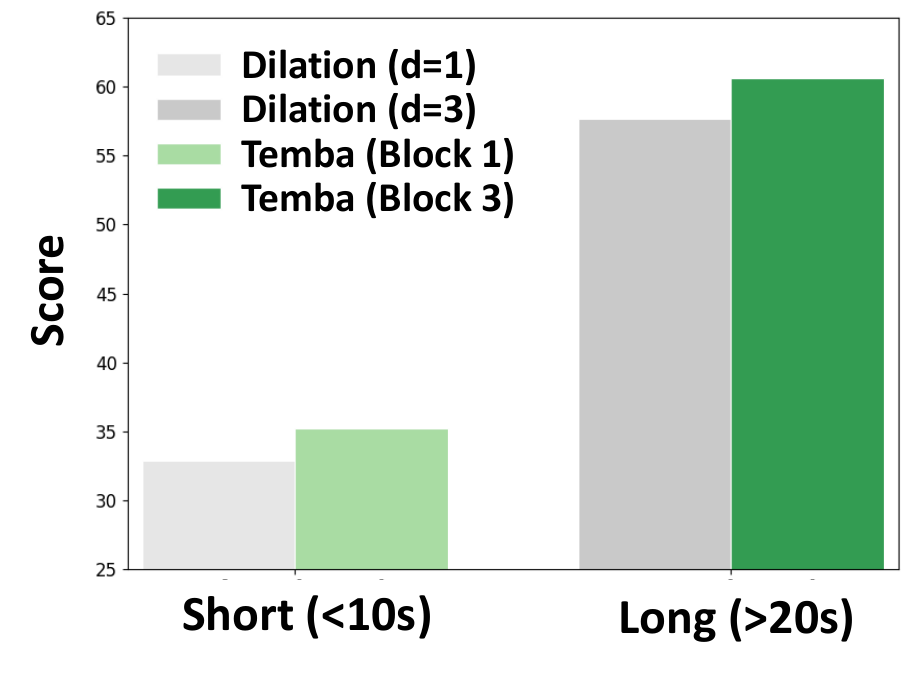}\vspace{-0.1in}
  \captionof{figure}{Impact of dilation in \block~for short and long actions}
  \label{fig:long_short} \vspace{-2mm}
\end{minipage}
\hfill
\begin{minipage}[t]{0.34\textwidth}
  \vspace{6pt}
  \centering
  \captionof{table}{Impact of each component of \modelname}
  \label{tab:ablation}
  \scalebox{0.72}{
  \begin{tabular}{ccc|ccc}
    \toprule
    \textbf{Temporal} & \textbf{Dilated} & \textbf{\fuserabbr} &
    \textbf{TSU} & \textbf{Charades} \\
    \textbf{Encoder} & \textbf{SSM} & & \textbf{(mAP)} & \textbf{(mAP)} \\
    \midrule
    \xmark & \xmark & \xmark & 24.7 & 22.9 \\
    $\checkmark$ & \xmark & \xmark & 40.2 & 32.4 \\
    $\checkmark$ & $\checkmark$ & \xmark & 42.5 & 32.5 \\
    \rowcolor{lightblue}
    $\checkmark$ & $\checkmark$ & $\checkmark$ &
    44.0 & 33.6 \\
    \bottomrule
  \end{tabular}} \vspace{-2mm}
\end{minipage}
\hfill
\begin{minipage}[t]{0.34\textwidth}
  \vspace{6pt}
  \centering
  \captionof{table}{Ablation on Loss components in \modelname}
  \label{tab:losses}
  \scalebox{0.72}{
  \begin{tabular}{cc|ccc}
    \toprule
    $\boldsymbol{\mathcal{L}_{\text{cons}}}$ &
    $\boldsymbol{\mathcal{L}_{\text{aux}}}$ &
    \textbf{TSU} & \textbf{Charades} & \textbf{Average}\\
     & & \textbf{(mAP)} & \textbf{(mAP)} & \textbf{(mAP)}\\
    \midrule
    \xmark & \xmark & 42.1 & 33.6 & 41.9\\
    $\checkmark$ & \xmark & 42.1 & 33.7 & 41.9\\
    \xmark & $\checkmark$ & 43.0 & 33.1 & 42.8\\
    \rowcolor{lightblue}
    $\checkmark$ & $\checkmark$ &
    44.0 & 33.6 & 43.8\\
    \bottomrule
  \end{tabular}}
\end{minipage} \vspace{-2mm}
\end{table*}
\vspace{-4mm}
\section{Experiments}
\label{sec:exp}
\textbf{Datasets.} 
We evaluate \modelname, on two challenging, real-world, multi-label action detection datasets: Toyota Smarthome Untrimmed~\cite{Dai_2022_PAMI}, and Charades~\cite{charades}. The datasets contain densely labeled activities, making the task of action detection challenging.
Toyota Smarthome Untrimmed (TSU)~\cite{Dai_2022_PAMI} is an untrimmed ADL video dataset capturing indoor activities related to daily living. It includes 51 action classes, with up to 5 co-occurring actions in a single frame, as well as composite actions. The average duration of the videos in TSU is 21 minutes, considerably higher than the other datasets making it challenging for action detection.
Charades~\cite{charades} is another large-scale dataset consisting of 9848 untrimmed videos spanning 157 actions. It contains 66500 annotations of daily activities with an average of 6.8 action instances per video.  We evaluate our methods on these datasets using the `\textit{action localization}' protocol described in~\cite{Dai_2022_PAMI}, and ~\cite{sigurdsson2017asynchronous}, respectively, to detect frame-level actions and compare them with the temporal annotations.

\noindent
\textbf{Implementation Details.}
In \modelname, we use I3D~\cite{i3d} and CLIP-L/14~\cite{CLIP} as our visual backbones. Their feature dimensions $D$ are 1024 and 768 respectively. Input features are initially transformed to an embedding of dimension $D_0 =256$ before being passed to the first Temba block. \modelname~consists of $K=3$ Temba blocks with a feature expansion ratio of $\gamma=1.5$. We follow the Mamba configurations of~\cite{vim}, with a state-dimension of 16. The dilation parameters are set to $\eta = k$ for Temba block $k$. 
For TSU, and Charades, the input Temporal dimensions are fixed at 2500, and  256, respectively. We set  $\alpha=100.0$ and $\beta=1$ in all experiments. In the Supplementary material, we provide configuration details for each of the datasets. We use Adam optimizer~\cite{adam_optimizer} and cosine learning rate scheduler. All our models are trained on a single $24$GB RTX A5000 GPU. 

\subsection{Comparison with State-of-the-Art}
In \Cref{tab:sota}, we benchmark \modelname~against the state-of-the-art Temporal Action Detection methods on TSU and Charades datasets. The top section demonstrates the performance of models using an I3D visual backbone for feature extraction which is predominantly used in previous works~\cite{TGM1, PDAN, MLAD, dai2021ctrn, dai2022mstct}. In the TSU dataset, while other methods struggle to detect actions in the videos due to their longer duration compared to other datasets, \modelname~outperforms the state-of-the-art methods, demonstrating its ability to learn effective long-range representations in the state space, directly tackling \textbf{\textcolor{challenge1}{Challenge 1}}. We observe that \modelname~achieves comparable performance with state-of-the-art~\cite{dai2022mstct} in Charades with $\mathbf{80\%}$ less parameters. 

The bottom section of \Cref{tab:sota}, presents the performance of temporal models with foundation models as the visual backbone. TTM~\cite{ryoo2023token} uses a ViViT~\cite{vivit} backbone, pretrained on the large-scale datasets JFT~\cite{jft_300m} and Kinetics-400~\cite{kinetics} and finetuned on the target dataset, for feature extraction whereas the other models use the pretrained foundation model CLIP~\cite{clip_representation}. 
We observe that \modelname~trained using CLIP features achieves state-of-the-art performance on both datasets and even outperforms \modelname~trained with I3D features on both TSU and Charades, highlighting the fact that foundation models learn more generalizable discriminative features compared to specialized models.  

In \Cref{fig:long_short}, we evaluate adaptive dilation in \modelname\ by comparing each \block~to its fixed-dilation counterpart. \block~Block 1 improves mAP on short actions ($<\!10$\,s), while \block~Block 3 excels on long actions ($>\!20$\,s). Thus, progressively increasing dilation across blocks enables \modelname~to effectively capture both fine-grained short-term dynamics and long-range temporal dependencies in untrimmed videos aligning with the goal of \textbf{\textcolor{challenge23}{Challenge 2}}. 

Furthermore, in~\Cref{tab:acmetrics}, we measure the \modelname's ability to model temporal dependencies and co-occurrence dependencies of actions on the TSU dataset, using the Action Conditioned Metrics introduced in~\cite{MLAD}. The metrics compute Precision ($P_{AC}$), Recall ($R_{AC}$), F1-Score ($F1_{AC}$) and mAP ($mAP_{AC}$) which are conditioned on actions occurring within a temporal window $\tau$. In~\Cref{tab:acmetrics}, for $\tau=0$ we observe that \modelname~achieves state-of-the-art performance indicating its effectiveness in detecting densely distributed overlapping actions thus addressing \textbf{\textcolor{challenge3}{Challenge 3}}. Furthermore, \modelname~outperforms the baselines for $\tau>0$ reaffirming \modelname's ability in capturing long-range dependencies between action instances. \vspace{-1mm}
\subsection{Ablation Study}
\vspace{-1mm}
\noindent
\textbf{Impact of each component in \modelname.} \Cref{tab:ablation} summarizes component-wise contributions. The baseline uses a simple classification head over CLIP features without temporal modeling. Adding a multi-layer Mamba temporal encoder~\cite{vim} yields a large gain, highlighting the value of temporal modeling in long untrimmed videos. Replacing standard scanning with \emph{Dilated Scanning} in SSMs improves TSU by \(+2.3\%\) and maintains performance on Charades, capturing actions with varied durations. The \fuserabbr~further boosts results by aggregating complementary cues across \modelname~blocks. 
Together, these components deliver consistent, significant improvements over the CLIP baseline, showing the benefit of each module of \modelname. \\
\noindent
\textbf{Impact of Loss Components.}
Following previous work, the baseline model is trained using only the Binary Cross-Entropy (\(\mathcal{L}_{\text{BCE}}\)) classification loss for action detection. As shown in \Cref{tab:losses}, incorporating the proposed consistency loss (\(\mathcal{L}_{\text{cons}}\)) alone yields minimal improvement, as it primarily acts as a regularizer on the projection space of SSM outputs ($C_i$ of i-th SSM in a block) without directly influencing the final prediction confidence. Additionally, introducing the scale-aware auxiliary loss (\(\mathcal{L}_{\text{aux}}\)) leads to a clear performance gain by encouraging intermediate \block~blocks to learn discriminative, scale-specific temporal representations. The best performance is achieved when both losses are combined with \(\mathcal{L}_{\text{BCE}}\), providing a $+1.9\%$ increase in average mAP across datasets. The proposed combination of losses is therefore instrumental in promoting coherent temporal alignment across dilated SSMs and enhancing multi-scale representation learning within \modelname. 
\vspace{-2.5mm}
\begin{table}[h]
\begin{minipage}{0.48\linewidth}
\centering
\caption{Impact of alignment objectives among dilated SSMs in \block~blocks}
\scalebox{0.7}{
\begin{tabular}{l|cc}
\toprule
\textbf{Loss}  & \textbf{TSU} & \textbf{Ch}\\
\midrule
No Loss   & 43.0 & 33.1 \\
Correlation Loss  & 43.5 & 33.1 \\
\rowcolor{lightblue} 
Consistency Loss & 44.0 & 33.6 \\
\bottomrule
\end{tabular}}  
\label{tab:consistency_loss}
\end{minipage}%
\hfill
\begin{minipage}{0.48\linewidth}
\centering
\caption{Analysis of multi-scale fusion strategies}
\scalebox{0.6}{
\begin{tabular}{l|cc}
\toprule
\textbf{Method} & \textbf{TSU} & \textbf{Ch} \\
\midrule
\block~Block 3 & 41.8 & 32.2 \\
\midrule
Concat + Proj. & 42.2  & 32.8 \\
Concat + Proj. + SSM & 42.4  & 32.4  \\
\midrule
Sum + Proj. & 42.5 & 32.5  \\
\rowcolor{lightblue} Sum + Proj. + SSM & 44.0 & 33.6  \\
\bottomrule
\end{tabular}}
\label{tab:fuser}
\end{minipage}
\end{table}
\subsection{Design Choices}
\noindent 
\textbf{Choice of Projection Alignment in \block~Blocks.}
\Cref{tab:consistency_loss} ablates coherence among dilated SSMs within each \block~block ($\eta \geq 2$).
The \emph{No Loss} baseline trains SSMs independently, without any constraint on their projected subspaces. Adding a correlation regularizer that encourages linear co-variation between projection matrices $\mathbf{C}_i$ and $\mathbf{C}_j$ yields a modest gain. Our \emph{consistency loss}, which maximizes pairwise cosine similarity between \emph{normalized} projection matrices, delivers the best and most stable improvements on both datasets. This indicates that aligning projection subspaces across parallel SSMs promotes coherent temporal semantics while preserving complementary scale-specific cues for temporal action detection. \\
\noindent
\begin{minipage}[t]{0.48\linewidth}
  \vspace{0pt}\centering
    \includegraphics[width=\linewidth]{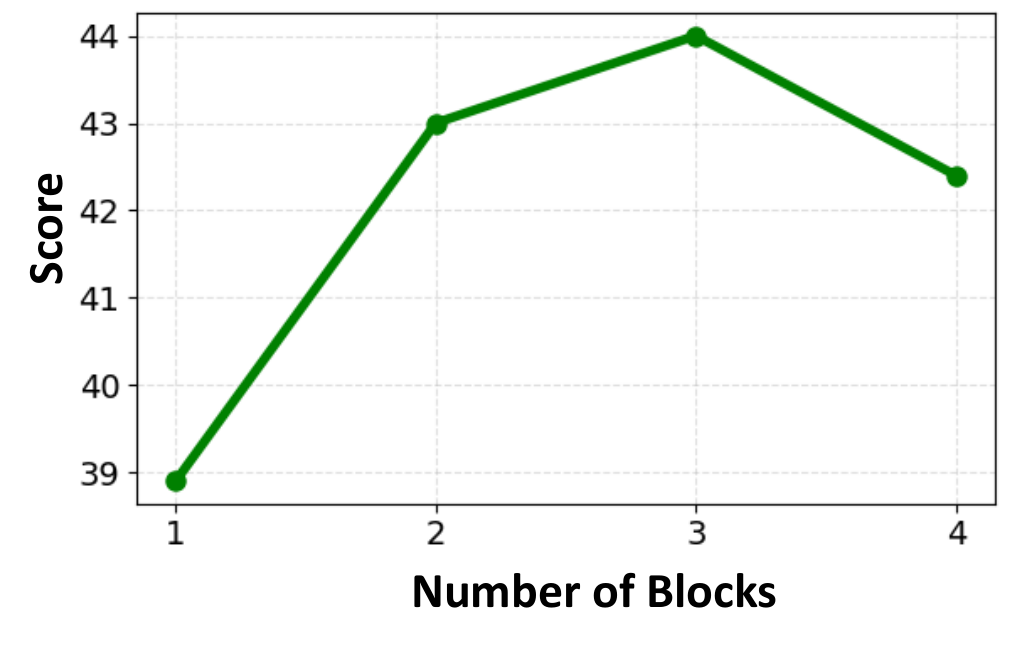}
    \captionof{figure}{Analysis of Number of \block~blocks}
    \label{fig:num_blocks} \vspace{2mm}
\end{minipage}\hfill
\begin{minipage}[t]{0.48\linewidth}
  \vspace{0pt}\centering
    \includegraphics[width=\linewidth]{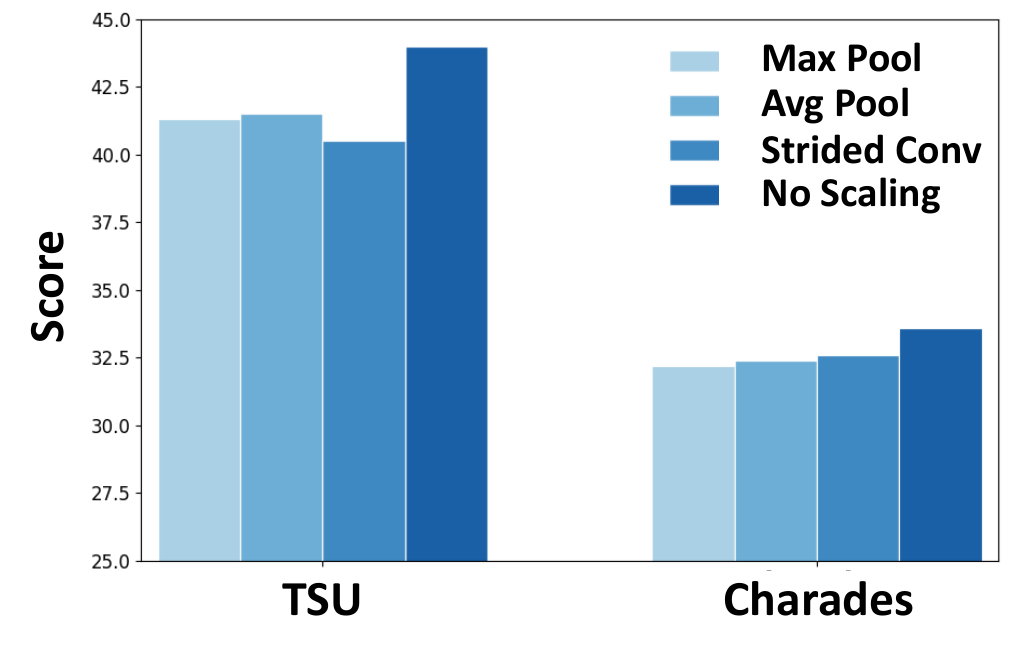}
    \captionof{figure}{Impact of Pooling Temporal Dimension}
    \label{fig:pool_temp} \vspace{2mm}
\end{minipage}
\noindent
\textbf{Number of Temba Blocks.} As shown in \Cref{fig:num_blocks}, increasing the number of stacked \block~blocks from one to three steadily improves performance, indicating that progressively larger dilations help capture richer temporal hierarchies. However, adding a fourth block causes a drop in accuracy, suggesting diminishing returns from deeper stacking. A three-block configuration achieves the best trade-off between temporal coverage and model effectiveness. \\
\noindent
\textbf{Impact of scaling temporal dimension in \block~Blocks.}
Prior work handles untrimmed videos at multiple temporal scales by first downsampling the temporal dimension and later interpolating them back to the input resolution. We evaluate analogous temporal scaling strategies within each \block~block. As shown in \Cref{fig:pool_temp}, max/avg pooling markedly degrades performance, indicating loss of fine-grained cues; strided convolutions also underperform due to temporal discontinuities. \modelname~performs best with no explicit scaling, allowing each block to operate at native resolution, preserving subtle transitions while capturing long-range dependencies in untrimmed videos. \\
\noindent
\textbf{Analysis of fusion strategies.}
\Cref{tab:fuser} compares multi-scale fusion in the \fuserabbr. Concatenating block outputs with a projection modestly outperforms the final \block~output, and adding an SSM on top yields limited extra benefit. In contrast, summation-based fusion better aligns features across scales, and an additional SSM further strengthens long-range temporal modeling. This combination attains the highest performance on both datasets, indicating that additive fusion preserves complementary cross-block information while the SSM refines the fused representation through temporal integration.\vspace{-0.5mm}
\vspace{-1.9mm}
\section{Beyond TAD: Video Summarization}
\label{sec:summarization}
\vspace{-1mm}
In addition to TAD, we demonstrate the generality of \modelname\ on the task of video summarization (\Cref{tab:summarize}).
This task condenses untrimmed videos into concise yet comprehensive summaries of key events and requires reasoning over both short-term visual changes and long-range temporal dependencies to estimate frame-level importance.
\begin{table}[h]
  \caption{Performance of \modelname~on Video Summarization}
  \centering
  \label{tab:summarize}
  \scalebox{0.85}{
  \setlength{\tabcolsep}{5pt}
  \begin{tabular}{lcccc}
    \toprule
    & \multicolumn{2}{c}{\textbf{SumMe}} & \multicolumn{2}{c}{\textbf{TVSum}} \\
    \cmidrule(lr){2-3}\cmidrule(lr){4-5}
    \textbf{Method} & $\tau$ & $\rho$ & $\tau$ & $\rho$ \\
    \midrule
    Random                & 0.000 & 0.000 & 0.000 & 0.000 \\
    Human                 & 0.205 & 0.213 & 0.177 & 0.204 \\
    \midrule
    \rowcolor{black!7} 
    \multicolumn{5}{l}{\emph{Visual only}} \\
    dppLSTM\, \cite{dpplstm}              & 0.040 & 0.049 & 0.042 & 0.055 \\
    DAC\,\cite{dac}    & 0.063 & 0.059 & 0.058 & 0.065 \\
    HSA\text{-}RNN\,\cite{hsarnn}        & 0.064 & 0.066 & 0.082 & 0.088 \\
    DAN\,\cite{dan}   & --    & --    & 0.071 & 0.099 \\
    STVT\,\cite{stvt}  & --    & --    & 0.100 & 0.131 \\
    DSNet\text{-}AF\,\cite{dsnet} & 0.037 & 0.046 & 0.113 & 0.138 \\
    DSNet\text{-}AB\,\cite{dsnet} & 0.051 & 0.059 & 0.108 & 0.129 \\
    VJMHT\,\cite{vjmht}  & 0.106 & 0.108 & 0.097 & 0.105 \\
    VASNet\,\cite{fajtl_vasnet_2019}  & 0.160 & 0.170 & 0.160 & 0.170 \\
    DMASum\, \cite{Wang_2020}& 0.063 & 0.089 & 0.203 & 0.267 \\
    RR\text{-}STG\,\cite{RR-STG} & 0.211 & 0.234 & 0.162 & 0.212 \\
    AAAM\,\cite{maam}   & --    & --    & 0.169 & 0.223 \\
    MAAM\,\cite{maam}   & --    & --    & 0.179 & 0.236 \\
    VSS\text{-}Net\,\cite{vssnet} & --    & --    & 0.190 & 0.249 \\
    CSTA\cite{csta} & 0.246 & 0.274 & 0.194 & 0.255 \\
    \midrule
    \rowcolor{black!7} 
    \multicolumn{5}{l}{\emph{Visual + Text}} \\
    CLIP\text{-}It\,\cite{clipit} & --    & --    & 0.108 & 0.147 \\
    iPTNet\,\cite{iptnet} & 0.101 & 0.119 & 0.134 & 0.163 \\
    A2Summ\,\cite{he_a2summ_2023} & 0.108 & 0.129 & 0.137 & 0.165 \\
    MSVA\,\cite{ghauri_msva_2021}    & 0.200 & 0.230 & 0.190 & 0.210 \\
    SSPVS\,\cite{haopeng_sspvs_2022}  & 0.192 & 0.257 & 0.181 & 0.238 \\
    LLMVS\,\cite{llmvs} & 0.253 & 0.282 & 0.211 & 0.275 \\
    \hline
    CLIP (w/o \modelname)\, & 0.197 & 0.221 & 0.207 & 0.270 \\
\rowcolor{lightblue} \textbf{\modelname}\, & \textbf{0.254} & \textbf{0.284} & \textbf{0.221} & \textbf{0.289} \\
    \bottomrule
  \end{tabular}} \vspace{-1mm}
\end{table}

\noindent\textbf{Video Summarization.} Video Summarization techniques typically model dependencies between frames of a video using LSTM  or Transformers \cite{Wang_2020}. VAS-Net~\cite{fajtl_vasnet_2019} uses soft self-attention, while DMA-SUM employs Self-attention twice to capture second-order changes. RR-STG~\cite{RR-STG} captures dependencies by modeling spatial association between frames based on objects in the scene. CSTA~\cite{csta}, rearranges frame-features into an image-like grid and employs a 2D-CNN to model the attention map and yield a spatio-temporal representation. Multi-modal approaches~\cite{clipit, he_a2summ_2023, ghauri_msva_2021, haopeng_sspvs_2022, llmvs} integrate textual features in addition to frame representations to enhance the feature space with complementary information. CLIP-It~\cite{clipit} generates frame-level captions from the videos and uses a transformer to model the representations after cross-attention between the video and text features. A2Summ~\cite{he_a2summ_2023} models temporal correspondence between the textual captions or transcript and visual features. LLMVS~\cite{llmvs} first generates frame-level captions using a Multimodal LLM and subsequently queries an LLM to produce frame-level importance scores based on the captions. A visual summary is created after processing the  latent features of the LLM using a Transformer. We model short and long term dependencies between visual-text features using dilated SSMs in \modelname. 

\noindent
\textbf{Dataset Details.} For Video Summarization, \modelname~is evaluated on standard benchmarks SumMe~\cite{summe} and TVSum~\cite{tvsum}. SumME is composed of 
25 videos recorded using egocentric cameras, both static and dynamic, on holidays, events, etc. The videos range from 30s to 6 mins with a mean duration of 2min 26s. TVSum consists of 50 videos from 10 genres comprising documentaries, vlogs, etc. The videos are 1 to 10 mins long with a mean duration of 3min 55s. The videos are annotated by 15 to 20 individuals.

\noindent
\textbf{Evaluation Metrics.} Following~\cite{summary_eval}, we evaluate the Video Summarization performance of \modelname~using correlation coefficients, Kendalls's $\tau$~\cite{kendall1945treatment} and Spearman's $\rho$~\cite{spearman_rho}. These rank correlation coefficients measure the similarities between the ground-truth scores and the predictions. 

\noindent
\textbf{Method.}
Following Clip-It~\cite{clipit}, we generate frame-level descriptions for uniformly sampled frames and extract CLIP features~\cite{CLIP}. We form video features by stacking visual CLIP embeddings and apply cross-attention with the text descriptors; the cross-attended sequence is then processed by \modelname. The classification head is replaced with a linear regression head~\cite{csta} to predict per-frame importance.

We evaluate the performance of the cross-attended input features (w/o \modelname) and \modelname~on Video Summarization on the TVSum~\cite{tvsum} and SumMe~\cite{summe} benchmarks (details in Supplementary). On both the datasets, \modelname~attains state-of-the-art results. We attribute the gains to the dilated SSMs within each \block, which enable modeling of temporal dependencies over varying time horizons, capturing both fine-grained transitions and long-range semantic structures essential for effective summarization. These findings reinforce the versatility of \modelname~as a generic framework for long-video understanding. 
\vspace{-2.3mm}
\section{Conclusion}
\label{sec:conc}
This work presents \modelname, a multi-scale Mamba-based approach for TAD in Untrimmed ADL Videos. By integrating dilation into the SSM formulation, and guiding each temporal scale through alignment and targeted supervision, \modelname~learns short-term atomic actions as well as extended activities. 
The \fuser~learns a robust unified representation from scale-specific representations, enabling accurate localization in densely annotated ADL settings. \modelname~achieves state-of-the-art performance on TSU and Charades setting a new benchmark for TAD in Activities of Daily Living with only 17M parameters, significantly outperforming existing methods. Beyond action detection, \modelname~also generalizes to Video Summarization, achieving state-of-the-art in standard benchmarks. This establishes \modelname~as a powerful and versatile framework for long-video understanding. 
\section*{Acknowledgements}
This work was supported in part by the National Science Foundation (IIS-2245652) and the University of North Carolina at Charlotte. 
\vspace{-6mm}
{
    \small
    \bibliographystyle{ieeenat_fullname}
    \bibliography{main}
}

\clearpage
\maketitlesupplementary

\appendix

\section*{Overview}
The supplementary is categorized into the following parts: \\
\begin{itemize}
    \item Section \ref{sec:config}: Training configurations for each Dataset
    \item Section \ref{sec:fuser-abl}: Effect of \fuserabbr~in \modelname
    \item Section \ref{sec:llm_comparison}: Limitations of MLLMs in TAD
    \item Section \ref{sec:flops}: Computational Footprint of \modelname
    \item Section \ref{sec:qual}: Qualitative Examples
    \item Section \ref{sec:temba_arch}: Architecture of \modelname
\end{itemize}

\section{Training configurations for each Dataset}
\label{sec:config}
\begin{table}[h] \vspace{-0.17in}
\centering
\caption{\textbf{\modelname~Training Settings}}
\scalebox{0.79}{
\begin{tabular}{c|c|c}
\toprule
 \textbf{Configuration} & \textbf{TSU} & \textbf{Charades}  \\
\midrule
Padded Temporal Length & $2500$  & $256$ \\ 
Optimizer & $AdamW$  & $AdamW$ \\ 
Momentum & $0.9$  & $0.9$ \\ 
Learning Rate (LR) & $4.5e^{-4}$  & $2.5e^{-4}$ \\ 
LR Scehduler & $Cosine$ &  $Cosine$ \\ 
Warmup Epochs & $5$ &  $5$ \\ 
Batch Size & $1$ & $5$ \\ 
Total Epochs & $140$ & $30$ \\ 
\bottomrule
\end{tabular}}  
\label{tab:config}
\end{table}


For all experiments, we pad the features of each input video to a fixed sequence length. \Cref{tab:config} shows the configurations used for each dataset.  




\section{Effect of \fuserabbr~in \modelname}
\label{sec:fuser-abl}
\begin{figure}[h!] \vspace{-0.14in}
    \centering
    \scalebox{0.6}{
        \includegraphics[width=0.5\textwidth]{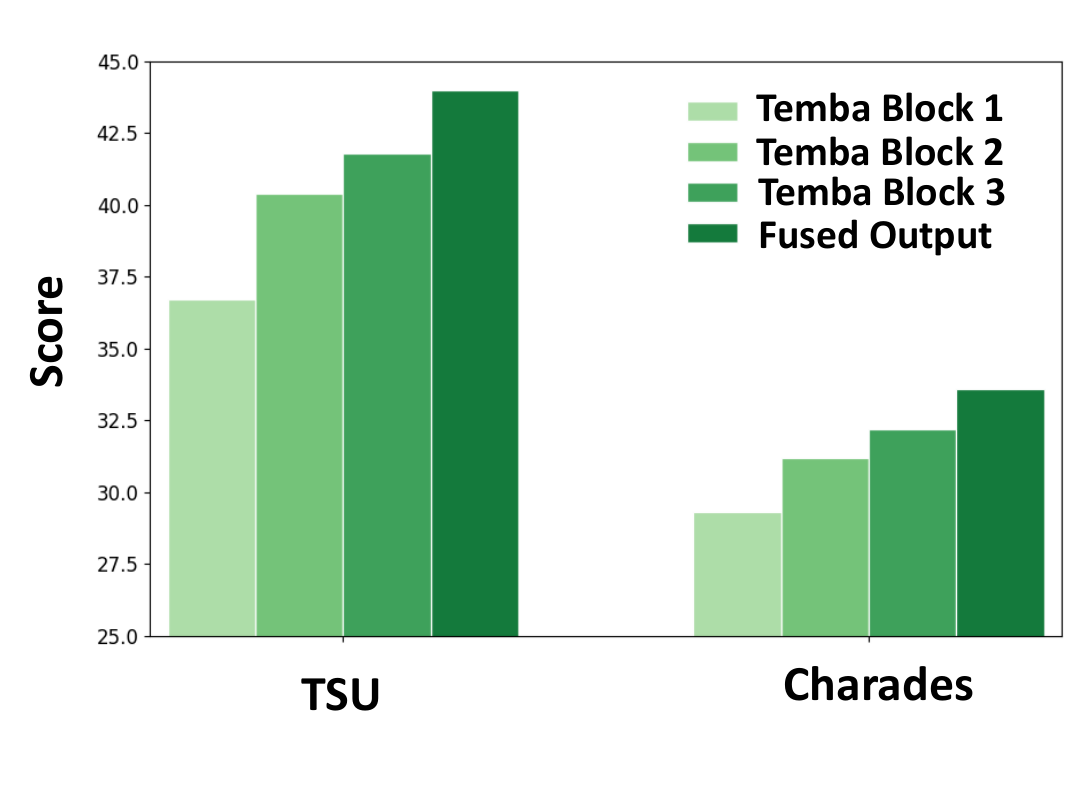}
    } \vspace{-0.17in}
    \caption{Impact of \fuser~in \modelname.}
    \vspace{-0.17in}
    \label{fig:fuser_abl}
\end{figure}
In~\Cref{fig:fuser_abl}, we illustrate the impact of \fuser~in effectively aggregating complementary information from the \block~blocks. As the temporal hierarchy deepens, later \block~blocks exhibit progressively higher performance, reflecting the refinement of temporal representations within \modelname. The \fuser~integrates these multi-scale outputs into a coherent state representation, enabling \modelname~to capture both fine-grained and long-range dependencies. This unified representation is central to the superior performance achieved by \modelname.

\section{Limitations of MLLMs in Temporal Action Detection}

\label{sec:llm_comparison}
\begin{figure}[h!]
    \centering
    \vspace{-0.25in}
    \scalebox{0.85}{
        \includegraphics[width=0.5\textwidth]{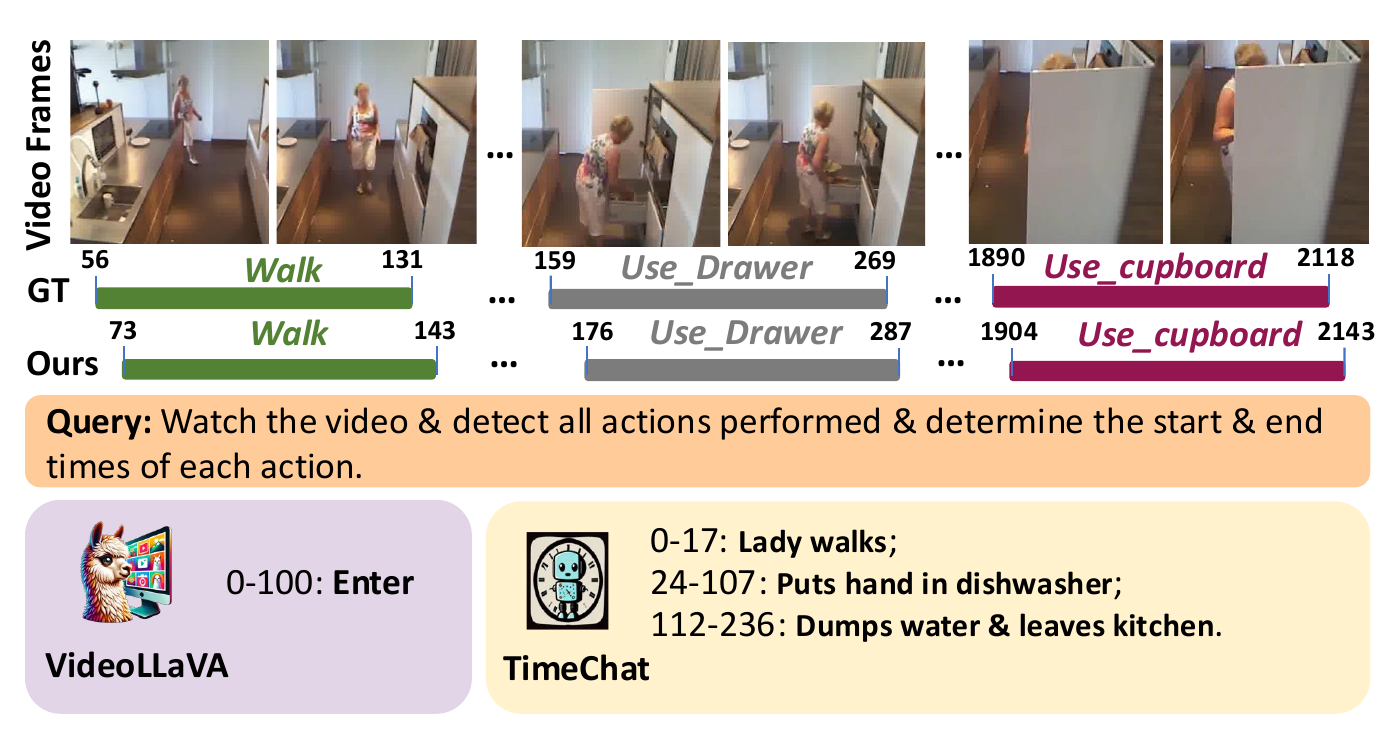}
    }
    \caption{Comparison of Multimodal LLMs (MLLMs) and our approach on long untrimmed videos.}
    \label{fig:LLM}
\end{figure}

While multimodal LLMs~\cite{videollava, videochatgpt} excel at capturing global spatio-temporal semantics for short clips, they often fail to identify the precise timestamps corresponding to frames of interest. TimeChat~\cite{timechat}, despite being trained on timestamps of input videos, fails to detect concurrent actions in complex videos. Furthermore, TAD in long videos using MLLMs requires enormous computational resources (7B parameters), making them infeasible for edge devices. As shown in Figure~\ref{fig:LLM}, both VideoLLaVa~\cite{videollava} and TimeChat~\cite{timechat} struggle to identify dense concurrent actions in untrimmed videos.

\section{Computational Footprint of \modelname}
\label{sec:flops}
\begin{table}[h] \vspace{-0.14in}
\centering
\caption{\textbf{\modelname~Computational Footprint}} 
\scalebox{0.79}{
\begin{tabular}{l|c|c}
\toprule
 \textbf{Method} & \textbf{FLOPs} $\downarrow$ & \textbf{Throughput} $\uparrow$  \\
 & (G)  & (Samples/sec) \\
\midrule
MS-TCT & 27.4 & 22.2\\ 
\modelname & \textbf{3.46}  &\textbf{51.0} \\ 
\bottomrule
\end{tabular}}  \vspace{-0.14in}
\label{tab:flops}
\end{table}
\modelname~demonstrates a substantially lighter computational footprint compared to the baseline MS-TCT~\cite{dai2022mstct}. As shown in~\Cref{tab:flops}, \modelname~requires only 3.46 GFLOPs nearly an order of magnitude lower than that of MS-TCT. This reduction in compute is also reflected in the runtime characteristics where \modelname~processes $51.0$ 
\begin{figure*}[h]
  \centering
  \scalebox{0.9}{
  \includegraphics[width=\linewidth]{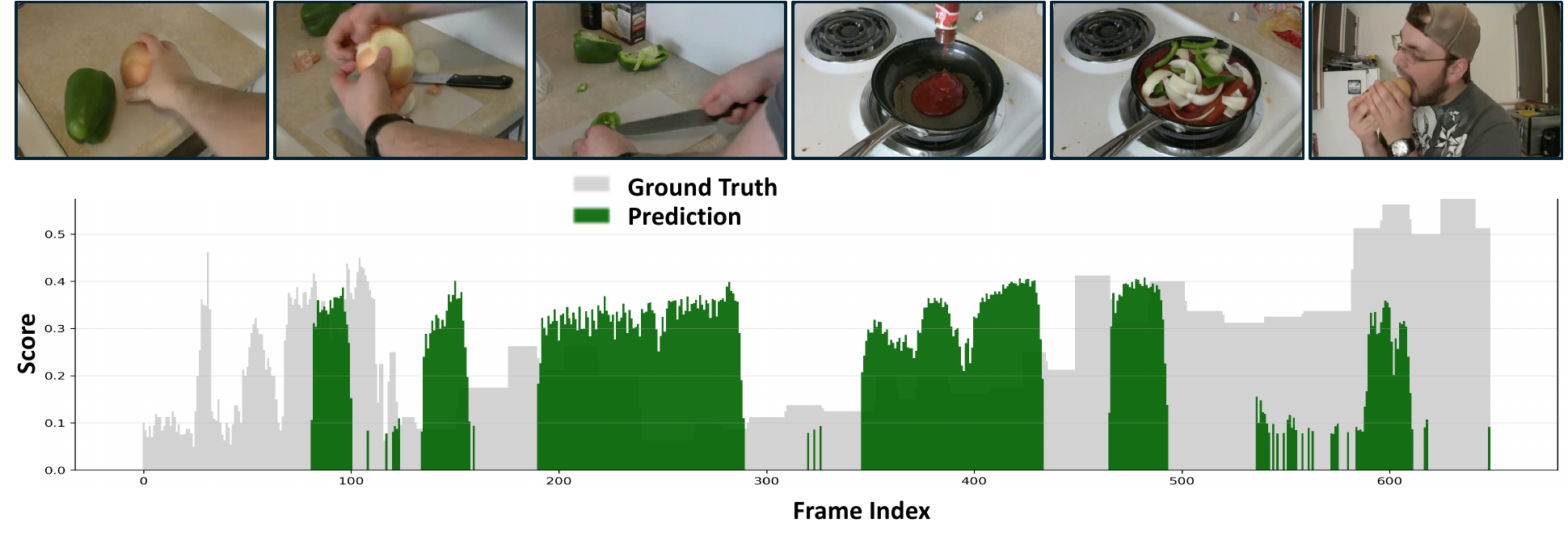}}
  \caption{\textbf{Qualitative Evaluation} of Video Summarization using \modelname~on an Example Video.}
  \label{fig:qualitative_summ}
\end{figure*}
%
\begin{table*}[h]
\centering
\caption{\textbf{\modelname~architecture}. The input and out feature size is following $T \times D$ format, where $T$ is the number of tokens and $D$ is the feature dimension. The hyper-parameters are defined as: $H$: linear hidden dimension, $D_s$: SSM dimension. }
\label{tab:architecture}
\scalebox{0.8}{
\begin{tabular}{c|l|l|c|c|c}
\hline
\textbf{Block}                                         & \textbf{Component}                      & \textbf{Learnable layer}     & \textbf{Hyper-parameters}             & \textbf{Input size}         & \textbf{Output size}        \\ \hline
Visual Backbone & Projection & Linear Layer & H: 256 & 2500$\times$1024 & 2500$\times$256 \\
\hline
\multicolumn{1}{c|}{\multirow{2}{*}{Temba Block 1}} & Projection                      & Linear Layer           & H: 256 & 2500$\times$256 & 2500$\times$256  \\ \cline{2-6} 
\multicolumn{1}{c|}{}                         & Dilated SSM                 & SSM Branch 1& $D_s$: 256             & 2500$\times$256  & 2500$\times$256  \\ \cline{2-6} 
\hline
\multicolumn{1}{c|}{\multirow{3}{*}{Temba Block 2}} & Projection                      & Linear Layer           & H: 384 & 2500$\times$256 & 2500$\times$384  \\ \cline{2-6} 
\multicolumn{1}{c|}{}                         & \multirow{2}{*}{Dilated SSM}                 & SSM Branch 1 & $D_s$: 384             & 2500$\times$384  & 2500$\times$384  \\ 
\multicolumn{1}{c|}{}                         &              & SSM Branch 2 & $D_s$: 384             & 2500$\times$384  & 2500$\times$384  \\ \cline{2-6} 
\hline
\multicolumn{1}{c|}{\multirow{4}{*}{Temba Block 3}} & Projection                      & Linear Layer           & H: 576 & 2500$\times$384 & 2500$\times$576  \\ \cline{2-6} 
                     & \multirow{3}{*}{Dilated SSM}                 & SSM Branch 1 & $D_s$: 576             & 2500$\times$576  & 2500$\times$576  \\ 
                    &                  & SSM Branch 2 & $D_s$: 576             & 2500$\times$576  & 2500$\times$576  \\ 
                        &                & SSM Branch 3 & $D_s$: 576             & 2500$\times$576  & 2500$\times$576  \\ \cline{2-6} 
\hline

\multicolumn{1}{c|}{\multirow{4}{*}{\fuserabbr}} &  \multirow{3}{*}{Projection}                      & Linear Layer           & H: 576 & 2500$\times$256 & 2500$\times$576  \\ 
&& Linear Layer           & H: 576 & 2500$\times$384 & 2500$\times$576  \\ 
&& Linear Layer           & H: 576 & 2500$\times$576 & 2500$\times$576  \\ 
\cline{2-6} 
\multicolumn{1}{c|}{}                         & Fuser                 & SSM & $D_s$: 576             & 2500$\times$576  & 2500$\times$576  \\ \cline{2-6} 
\hline
\end{tabular}}
\end{table*}
samples per second, more than $\mathbf{2 \times}$ of MS-TCT. These results show that \modelname~not only reduces the computational burden but also delivers markedly higher throughput, highlighting its suitability for processing long untrimmed videos.
%
\begin{figure}[h]
  \scalebox{1.0}{
  \includegraphics[width=\linewidth]{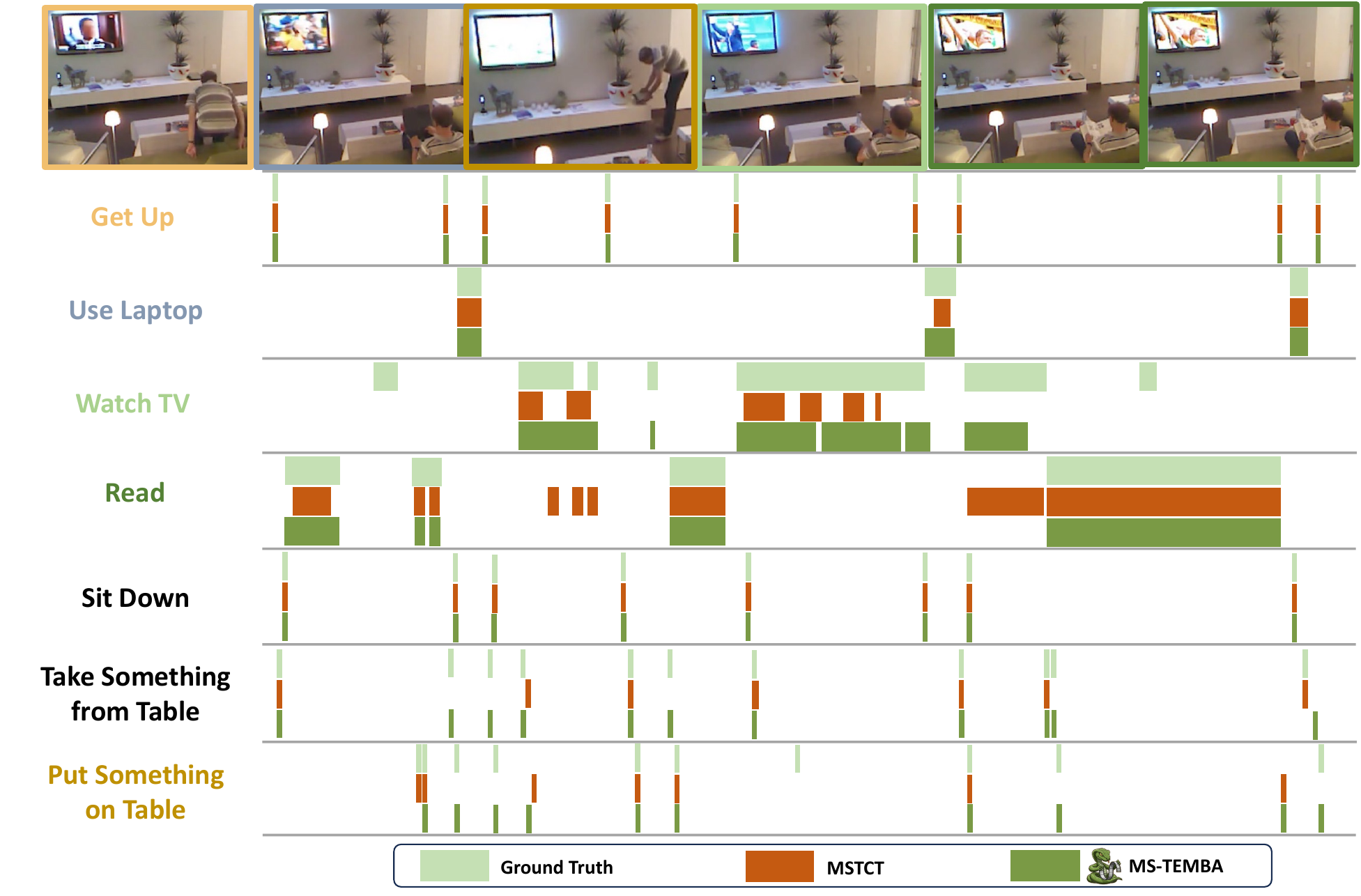}}
  \caption{\textbf{Qualitative Evaluation} of Temporal Action Detection on an Example Video.}
  \label{fig:qualitative}
\end{figure}
%
\section{Qualitative Examples.}
\label{sec:qual}
\noindent
\textbf{Video Summarization.} In \Cref{fig:qualitative_summ}, we present a qualitative example of video summarization using \modelname. The top row shows representative frames selected by the model, reflecting key events across the video. The bottom plot compares the predicted importance scores with the human-annotated ground truth. We observe that \modelname~assigns consistently higher scores to semantically meaningful segments, such as \textit{preparation of ingredients}, \textit{cooking transitions}, and \textit{final product}, while suppressing less informative regions. The close alignment between predicted and ground-truth importance curves highlights the model’s ability to capture both fine-grained and long-range temporal structure, resulting in coherent and human-aligned video summaries.

\noindent
\textbf{Temporal Action Detection.}
\Cref{fig:qualitative} presents a qualitative comparison of the predictions generated by \modelname~and MS-TCT on a sample untrimmed video. Notably, \modelname~demonstrates superior capability in detecting short-duration actions, such as \textit{Putting Something on a Table} and \textit{Take Something from Table}, some instances of which MS-TCT fails to recognize. For long-duration actions like \textit{Watch TV} or \textit{Read}, \modelname~predicts more accurate action boundaries and has fewer misclassifications.

\section{Architecture of \modelname}
\label{sec:temba_arch}
\Cref{tab:architecture} demonstrates the input and output feature size for each component of \modelname~. First, the Visual backbone extracts features of dimension $T \times D$ (where $D=1024$ for I3D backbone and $D=768$ for CLIP backbone). These features are projected using a Linear Projection layer to  $T \times 256$, which is the input to the first Temba block. Similar to TSU, here we fix the number of temporal segments as $T=2500$. Each \block~Block consists of a Projection Layer and Dilated SSM branches. The outputs of each Temba Block are aggregated in the \fuserabbr. In \Cref{tab:architecture}, we show the flow of these features in \modelname.

\end{document}